\documentclass[twoside,leqno,twocolumn]{article}

\usepackage[letterpaper]{geometry}

\usepackage{ltexpprt}
\usepackage{hyperref}

\usepackage{amsmath,amssymb}
\usepackage{graphicx}
\usepackage{booktabs}
\usepackage{tabularx}
\usepackage{multirow}

\usepackage{bm}
\usepackage{balance}
\usepackage{enumitem}
\setlist{leftmargin=3.5mm}

\begin{document}

\newcommand\relatedversion{}
\renewcommand\relatedversion{\thanks{The full version of the paper can be accessed at \protect\url{https://arxiv.org/abs/1902.09310}}} 

\title{\Large Heterogeneous Temporal Graph Neural Network}
\author{Yujie Fan$^1$
\and Mingxuan Ju$^{1,2}$
\and Chuxu Zhang$^3$
\and Liang Zhao$^4$
\and Yanfang Ye$^{* 1,2}$
}

\date{}

\maketitle







\begin{abstract} 
\small\baselineskip=9pt Graph neural networks (GNNs) have been broadly studied on dynamic graphs for their representation learning, majority of which focus on graphs with homogeneous structures in the spatial domain. However, many real-world graphs - i.e., \textit{heterogeneous temporal graphs} (HTGs) - evolve dynamically in the context of heterogeneous graph structures. The dynamics associated with heterogeneity have posed new challenges for HTG representation learning. To solve this problem, in this paper, we propose \textit{heterogeneous temporal graph neural network} (HTGNN) to integrate both spatial and temporal dependencies while preserving the heterogeneity to learn node representations over HTGs. Specifically, in each layer of HTGNN, we propose a hierarchical aggregation mechanism, including intra-relation, inter-relation, and across-time aggregations, to jointly model heterogeneous spatial dependencies and temporal dimensions. To retain the heterogeneity, intra-relation aggregation is first performed over each slice of HTG to attentively aggregate information of neighbors with the same type of relation, and then intra-relation aggregation is exploited to gather information over different types of relations; to handle temporal dependencies, across-time aggregation is conducted to exchange information across different graph slices over the HTG. The proposed HTGNN is a holistic framework tailored heterogeneity with evolution in time and space for HTG representation learning. Extensive experiments are conducted on the HTGs built from different real-world datasets and promising results demonstrate the outstanding performance of HTGNN by comparison with state-of-the-art baselines. Our built HTGs and code have been made publicly accessible at: \href{https://github.com/YesLab-Code/HTGNN}{https://github.com/YesLab-Code/HTGNN}.
\end{abstract}

\section{Introduction}
\footnotetext[1]{Case Western Reserve University, Cleveland, OH 44106, USA}
\footnotetext[2]{University of Notre Dame, Notre Dame, IN 46556, USA}
\footnotetext[3]{Brandeis University, Waltham, MA 02453, USA}
\footnotetext[4]{Emory University, Atlanta, GA 30322, USA \\ $^*$Corresponding author, Email: yanfang.ye@case.edu}

Many real-world data come in the form of graphs, such as academic networks \cite{hu2020open,wang2019heterogeneous}, social networks \cite{liao2018attributed,fan2019graph}, and epidemiological networks \cite{deng2020cola,kapoor2020examining}. The graph structure consists of a set of nodes interconnected by a set of edges. Learning node representations on graphs is essential for various downstream tasks, such as node classification, link prediction, and recommendation.

Recently, graph neural networks (GNNs) have been broadly studied and achieved state-of-the-art performance by taking both node features as well as graph structures into consideration. Despite their superior performance, most of the current research efforts concentrate on static graphs \cite{kipf2016semi,velivckovic2017graph,fu2020magnn,hu2020heterogeneous} or dynamic/spatial-temporal graphs with homogeneous structures \cite{yu2017spatio,sankar2020dysat,deng2020cola,nguyen2018continuous}. However, many real-world graphs evolve dynamically in the context of heterogeneous graph structures. From the perspective of spatial domain, the graph is heterogeneous with multi-typed nodes connected by multi-typed relations; from the perspective of temporal domain, either the node features or graph structures evolve over time. We call this type of graph the \textit{heterogeneous temporal graph} (HTG). A HTG could be described as an ordered list of heterogeneous graph slices with a set of temporal relations connecting them. It is a general concept for modeling heterogeneous and dynamically changing graph data. Typical examples include dynamic academic networks and epidemiological networks. For dynamic academic networks, the heterogeneous structures would evolve along with the authors' research directions and co-authorship. In contrast, the graph structures remains unchanged for epidemiological networks, but the node features inevitably change with increased/decreased patient numbers. It is worth noting that dynamic heterogeneous graphs \cite{yin2019dhne,xue2020modeling,luo2020dynamic,xie2021learning} can be treated as an instance of HTGs, where the dynamic nature comes from the evolving graph structures.

The dynamics associated with heterogeneity have posed new challenges for representation learning on HTGs. There exist some preliminary works. They could be roughly summarized into two categories: one first explores neural sequence models to process time-series features attached on each node, and then performs graph representation learning with the processed node features on the spatial domain \cite{wu2019graph,deng2020cola,kapoor2020examining,hong2020heteta}; the other first applies GNNs on each graph slice of a HTG, and then employs sequence models on the outputs of each slice to obtain the final representations \cite{yang2020dynamic,luo2020dynamic,xue2020modeling}. Although these works could achieve satisfactory results, they are still faced with the following limitations: (1) The existing models are graph-dependent. That is, the performance depends heavily on the characteristics of a graph. Specifically, for a HTG with dynamically evolving heterogeneous graph structures (e.g., academic networks), the second category approaches that emphasize more on spatial dependencies usually obtain better results. On the contrary, for a HTG with constantly changing node features (e.g., epidemiological networks), the first category methods that focus more on temporal dependencies would achieve superior performance. Apparently, selecting a model that best fits with a given HTG requires empirical knowledge. (2) The spatial and temporal dependencies are processed in a serialized way. Most existing models either analyze the temporal domain first and the spatial domain later or in reverse order, which weakens the spatial-temporal interactions as the information in these two domains is treated separately.

Currently, it is not yet well understood how to jointly integrate both spatial and temporal dependencies while preserving the heterogeneity for node representation learning over HTGs. To fill this gap, in this paper, we propose \textit{heterogeneous temporal graph neural network} (HTGNN), a holistic framework tailored heterogeneity with evolution in time and space to learn node representations on HTGs. More specifically, to retain the \textit{spatial heterogeneity}, we design intra-relation aggregation and inter-relation aggregation, which are performed purely on each graph slice of a HTG, to successively aggregate the information of a target node's neighbors within the same type of relation and over different types of relations. To handle the \textit{temporal dependencies}, we introduce across-time aggregation, which is conducted across different graph slices, to gather the information of the target node's temporal neighbors. To capture the \textit{spatial-temporal interactions}, we equip each layer of HTGNN with a hierarchical aggregation mechanism, including intra-relation, inter-relation, and across-time aggregation modules, to jointly, rather than serially, model heterogeneous spatial dependencies and temporal dimensions. With increased model depth, the information is iteratively propagated in these two domains, allowing HTGNN agnostic to graph characteristics. In sum, we make the following contributions:
\begin{itemize}[leftmargin=*]\setlength{\itemsep}{0pt}
	\item We study the representation learning problem on HTGs. HTG is a general concept to model graph data with heterogeneous spatial structures and temporal evolution patterns (i.e., dynamically evolving graph structures or constantly changing node features).
	\vspace{-0.05in}
	\item We propose HTGNN to learn node representations on HTGs. HTGNN is a holistic framework, which is capable of jointly modeling heterogeneous spatial dependencies and temporal dimensions. This character differs from existing works that process these two types of dependencies in a serialized way.
		\vspace{-0.05in}
	\item We establish two HTGs from different real-world datasets, one with dynamically evolving heterogeneous structures (i.e., OGBN-MAG) and another with constantly changing node features (i.e., COVID-19). Extensive experiments on these two HTGs demonstrate that HTGNN consistently achieves strong performance in comparison with state of the arts for different graph mining tasks.
\end{itemize}

\section{Related Work}

\noindent\textbf{Heterogeneous Graph Neural Networks.} Recently, various heterogeneous GNNs \cite{schlichtkrull2018modeling,zhang2019heterogeneous,hu2020heterogeneous,wang2019heterogeneous,fu2020magnn,zhao2021multi,zhao2021heterogeneous} have been proposed with successful applications \cite{liu2018heterogeneous,hou2021disentangled,ye2020alpha,fan2020metagraph,ye2020community}. RGCN \cite{schlichtkrull2018modeling} introduces relation-specific transformations for different relations during the learning process. HGT \cite{hu2020heterogeneous} utilizes meta relations to model graph heterogeneity and further learns the mutual attention for each meta relation based on the Transformer architecture. By leveraging metapaths defined on the heterogeneous graphs, HAN \cite{wang2019heterogeneous} designs node-level attention and semantic-level attention to learn the importance of metapath-based neighbors and different metapaths, respectively. These models are built for static heterogeneous graphs and cannot deal with the dynamic properties of HTGs. It is worth noting that, HGT uses a relative temporal encoding to assign each node a timestamp to handle graph dynamics. However, this could partially address the problem as the node embedding in HGT is assumed to be time-invariant, which is not in line with many real-world scenarios. 


\noindent\textbf{Dynamic Graph Learning.} Spatial-temporal graphs \cite{yu2017spatio,wu2019graph,kapoor2020examining,deng2020cola} and dynamic graphs with homogeneous structures \cite{sankar2020dysat,xu2020inductive,nguyen2018continuous} have been widely studied in the literature. To further consider the graph heterogeneity, learning on dynamic heterogeneous graphs has drawn increasing attention, including dynamic heterogeneous graph embedding models \cite{wang2020dynamic,peng2021lime,xie2021learning,yin2019dhne} that solely consider graph structures and dynamic heterogeneous GNNs \cite{xue2020modeling,yang2020dynamic,li2020heterogeneous,fan2021htg} that take both graph structure and node features into consideration. DyHATR \cite{xue2020modeling} first introduces node-level and edge-level attentions to learn heterogeneous information and then applies RNNs with temporal attention to capture temporal dependencies. HDGAN \cite{ji2021dynamic} combines heterogeneous attention and Hawkes process to model graph heterogeneity and dynamics. However, there are some limitations in current works. Firstly, these models are graph-dependent, which requires empirical knowledge; secondly, the spatial and temporal dependencies are processed serially, which leads to a weakened connection between these two domains. This paper addresses these issues by designing a holistic graph-agnostic framework to learn node representations on HTGs.


\section{Preliminary}\label{sec:preliminary}

In this section, we define concepts used in our model. 

\begin{Definition}
	\textbf{Heterogeneous Graph}. A heterogeneous graph is defined as a directed graph $G = (V, E)$ associated with a node type mapping $\phi$: $V \to A$ and a edge type mapping $\psi$: $E \to R$, where $V$ and $E$ denote the node set and edge set, $A$ and $R$ are the node type set and edge type set, with the constrain of $|A|+|R|>2$.
\end{Definition}

\begin{figure}[!h]
	\centering\vspace{-0.2cm}
	\includegraphics[width=0.95\linewidth]{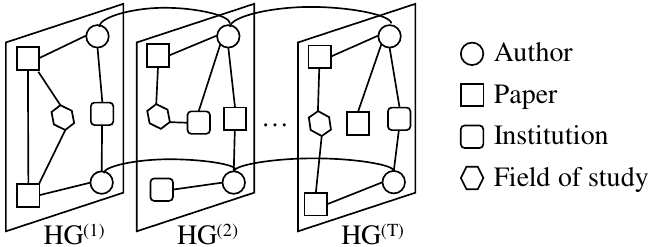}\vspace{-0.3cm}
	\caption{An graphical illustration of HTG.}\vspace{-0.3cm}
	\label{fig:exp}
\end{figure}

\begin{Definition}
	\textbf{Heterogeneous Temporal Graph.} A heterogeneous temporal graph is defined as a graph $\mathcal{G} = \big(\{G^{t}\}_{t=1}^T, E'\big) = (\mathcal{V}, \mathcal{E})$, where $T$ is the number of timestamps, $G^{(t)}$ is a heterogeneous graph at timestamp $t$, $E'$ describes the temporal relations between $G^{t}$ and $G^{t+1}$, $\mathcal{V} = \bigcup_{t=1}^T V^t$ and $\mathcal{E} = \bigcup_{t=1}^T E^t \cup E'$ denote the node set and edge set of $\mathcal{G}$, respectively. Figure \ref{fig:exp} shows one example from OGBN-MAG dataset. 
\end{Definition}

\begin{Definition}
	\textbf{Relation-$r$-based Neighbors.} Given a relation type $r\in R$, a node $v$ at timestamp $t$, its relation-$r$-based neighbors at timestamp $t$ is defined as $\mathcal{N}^t_{r}(v) = \{u| (u,v)\in E^{t}, \psi(u,v)=r \}$.
\end{Definition}

\begin{Definition}
	\textbf{Heterogeneous Temporal Graph Representation Learning.} Given a HTG $\mathcal{G} = \big(\{G^{t}\}_{t=1}^T, E'\big) = (\mathcal{V}, \mathcal{E})$ consisting of $T$ timestamps, the task of HTG representation learning is to learn a general $d$-dimensional node representation $\mathbf{h}_v\in \mathbb{R}^d$ for each node $v\in \mathcal{V}$ with $d\ll |\mathcal{V}|$. The node representations are able to capture both spatial heterogeneity and temporal dependencies of the HTG, and could be applied in various downstream tasks at timestamp $T+1$.
\end{Definition}

\begin{figure*}[t]
	\centering
	\includegraphics[width=0.98\linewidth]{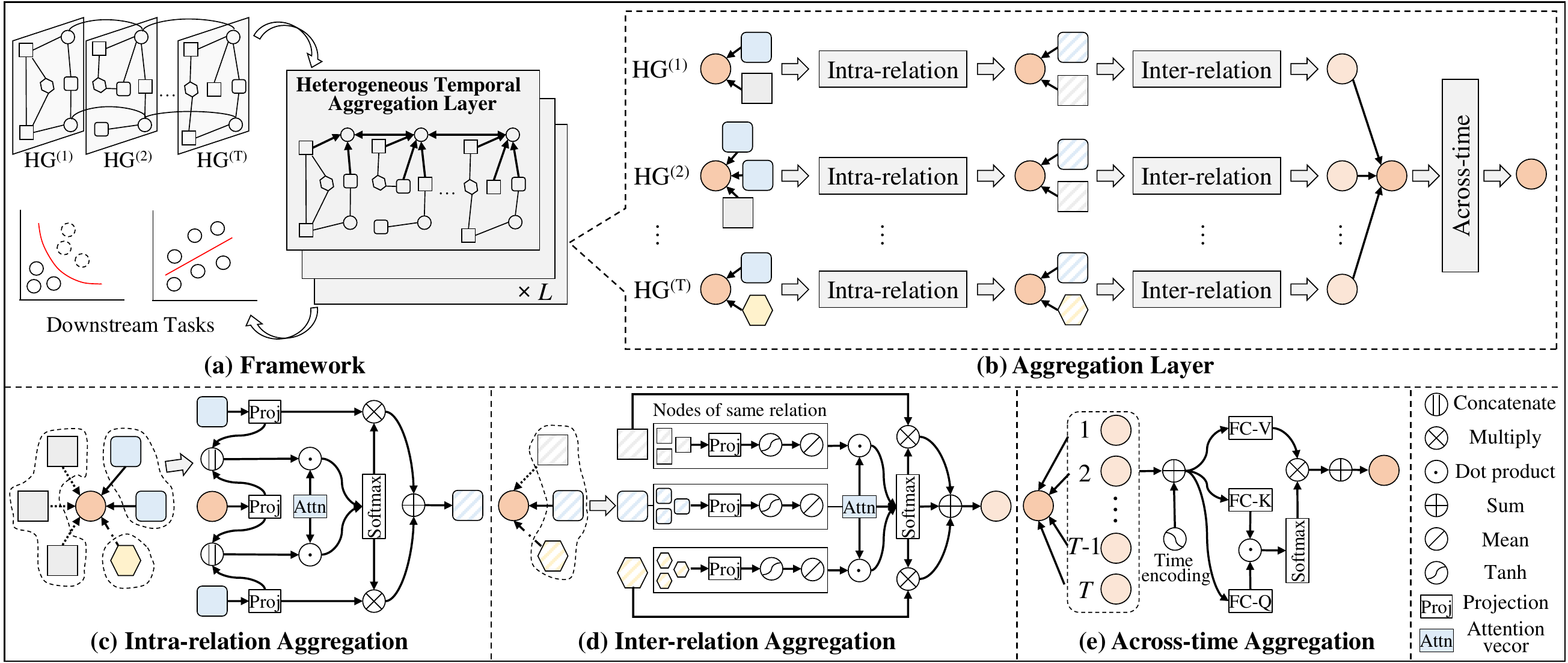}
	\vspace{-0.4cm}
	\caption{The overall framework of heterogeneous temporal graph neural network.}
	\label{fig:sys}\vspace{-0.5cm}
\end{figure*}

\section{Methodology}\label{sec:methodology}

\subsection{Overview}

The framework of HTGNN is shown in Figure \ref{fig:sys}. HTGNN takes a HTG as input and yields node representations as outputs for downstream tasks (see Figure \ref{fig:sys} (a)). It is composed of multiple heterogeneous temporal aggregation layers. Each layer is equipped with a hierarchical aggregation mechanism (see Figure \ref{fig:sys} (b)), including intra-relation, inter-relation, and across-time aggregation modules. Intra-relation and inter-relation aggregation modules are performed purely on each graph slice, aiming to depict the spatial heterogeneity, while the across-time aggregation module is conducted across different graph slices, aiming to capture the temporal dependencies. In one aggregation layer, each node successively receives messages from its spatial neighbors of the same relation type and different relation types; the nodes then start gathering messages from their temporal neighbors across graph slices. After this, another layer follows with node embeddings obtained from the previous layer. With increased model depth, the messages are iteratively propagated in spatial and temporal domains. Such design makes HTGNN be a holistic framework jointly modeling heterogeneous spatial dependencies and temporal dimensions. 

\subsection{Intra-relation Aggregation}

In a heterogeneous graph, each node type may have its own feature space. Take OGBN-MAG dataset as an example where only nodes with paper type are associated with input features. Metapath2vec~\cite{dong2017metapath2vec} is a common approach to initiate features for those nodes without input features. Apparently, the feature spaces for paper type and other types are different as the former reflects the text content information while the latter represents the graph structural information. To handle this problem, before feeding node features into multiple heterogeneous temporal aggregation layers, we first adopt a type-specific projection on each node to map its distinct feature vector into a same feature space. Mathematically, given a node $v$ with node type $\phi(v)$ at timestamp $t$, we have: 
\begin{equation}
	\mathbf{h}^{t,0}_{v} = \mathbf{W}_{\phi(v)}\cdot\mathbf{x}^t_v,
\end{equation}
where $\mathbf{x}^t_v\in \mathbb{R}^{d'}$ and $\mathbf{h}^{t,0}_v\in \mathbb{R}^{d}$ are $d'$-dimensional raw feature vector and $d$-dimensional projected embedding of node $v$, respectively; $\mathbf{W}_{\phi(v)}\in \mathbb{R}^{d\times d'}$ is the trainable type-specific transformation matrix.

The intra-relation aggregation module is performed separately on each relation type in each graph slice. It takes the node embeddings of last layer as inputs and outputs multiple relation embeddings for each node at each timestamp. Given a target node $v$ at timestamp $t$ and a relation type $r\in R$, the intra-relation aggregation can be described as:
\begin{equation}
	\mathbf{h}^{t,l}_{v,r} = AGG_{intra}\Big(\big\{\mathbf{h}^{t,l-1}_u|u\in \mathcal{N}^t_r(v)\big\};\boldsymbol{\Theta}_{intra}\Big),
\end{equation}
where $\mathcal{N}^t_r(v)$ represents relation-$r$-based neighbors of node $v$ at timestamp $t$, $\mathbf{h}^{t,l-1}_u$ is node $u$'s embedding at timestamp $t$ in layer $l-1$, $\mathbf{h}^{t,l}_{v,r}$ denotes the relation $r$'s embedding with respect to node $v$ at timestamp $t$ in layer $l$, and $\boldsymbol{\Theta}_{intra}$ is the trainable parameters and is non-shareable for relation type, timestamp and aggregation layer. Noted that when $l=1$, $AGG_{intra}$ takes the outputs of type-specific projection module introduced above as inputs.

For a target node, its different neighbors, even within the same type of relation, would also contribute differently in the learning process, and thus we adopt the self-attention mechanism to assign each neighbor a weight reflecting its importance. Formally, given a target node $v$ and one of its relation-$r$-based neighbors at timestamp $t$, i.e., $u\in \mathcal{N}^t_r(v)$, their attention coefficient can be computed by:
\begin{equation}
	e_{(u,v),r}^{t,l} = \sigma\Big(\big[\mathbf{a}_{r}^{t,l}\big]^\top \cdot \big[\mathbf{W}_{r}^{t,l}\cdot\mathbf{h}^{t,l-1}_u \| \mathbf{W}_{r}^{t,l}\cdot\mathbf{h}^{t,l-1}_v\big]\Big),
\end{equation}
where $\sigma(\cdot)$ is LeakyReLU, $\mathbf{W}_{r}^{t,l}\in \mathbb{R}^{d\times d}$ and $\mathbf{a}_{r}^{t,l}\in \mathbb{R}^{2d}$ are trainable transformation matrix and attention vector, respectively, and $\|$ concatenates the vector. We then normalize the attention coefficient across all relation-$r$-based neighbors via softmax function:
\begin{equation}
	\alpha_{(u,v),r}^{t,l} = \frac{\exp\Big(e_{(u,v),r}^{t,l}\Big)}{\sum_{u' \in \mathcal{N}^t_r(v)} \exp\Big(e_{(u',v),r}^{t,l}\Big)}.
\end{equation}
After obtaining the normalized attention values of node $v$'s neighbors, we perform weighted combination to compute relation $r$'s embedding with respect to $v$:
\begin{equation}
	\mathbf{h}^{t,l}_{v,r} = \sigma\bigg(\sum_{u\in \mathcal{N}^t_r(v)}\big[\alpha_{(u,v),r}^{t,l}\big] \cdot \big[\mathbf{W}_{r}^{t,l}\cdot\mathbf{h}^{t,l-1}_u\big]\bigg).
\end{equation}
Figure \ref{fig:sys} (c) illustrates the implementation of intra-relation aggregation module. Without loss of generality, multi-head attention mechanism can be employed. Specifically, $K$ independent attention heads are executed in a parallel fashion, and then their representations are concatenated, as following:
\begin{equation}
	\mathbf{h}^{t,l}_{v,r} = \mathop\|_{k=1}^K \sigma\bigg(\sum_{u\in \mathcal{N}^t_r(v)}\big[\alpha_{(u,v),r}^{t,l}\big]^k \cdot \big[\mathbf{W}_{r}^{t,l}\cdot\mathbf{h}^{t,l-1}_u\big]^k\bigg),
\end{equation}
where $\big[\alpha_{(u,v),r}^{t,l}\big]^k$ denotes the attention value at $k$-th attention head.

\subsection{Inter-relation Aggregation}

Through intra-relation aggregation, the target node would gather multiple relation embeddings. Based on this, the inter-relation aggregation module aims to learn a spatial embedding for the target node summarizing the information of its spatial neighbors over all relation types. Formally, this process is denoted as: 
\begin{equation}
	\mathbf{h}^{t,l}_{v,R} = AGG_{inter}\Big(\big\{\mathbf{h}^{t,l}_{v,r}|r\in R(v)\big\};\boldsymbol{\Theta}_{inter}\Big),
\end{equation}
where $R(v)$ denotes the set of relations connected to node $v$, $\mathbf{h}^{t,l}_{v,r}$ is relation $r$'s embedding with respect to node $v$ from previous module, $\mathbf{h}^{t,l}_{v,R}$ denotes the spatial embedding of node $v$ that will be learned in this module, $\boldsymbol{\Theta}_{inter}$ is the trainable parameters and is non-shareable for timestamp and aggregation layer. 

A straightforward way to implement $AGG_{inter}(\cdot)$ is treating each relation embedding equally by conducting element-wise sum/mean operation or concatenating them followed by a linear transformation. However, each relation type preserves a unique semantic meaning and thus should not be treated identically. Therefore, we manage to learn a importance weight for each relation type and explore the attention mechanism for implementation. Specifically, for relation type $r$, we use a three-step process to learn its importance: (1) we first retrieve its embeddings of all related nodes and feed them into a non-linear transformation; (2) we generate its summarized embedding by averaging the transformed relation embeddings; (3) finally, we calculate its attention coefficient by measuring the similarity between its summarized embedding with a relation attention vector. This learning process is formalized as:
\begin{equation}
	e^{t,l}_{r} = \big[\mathbf{c}^{t,l}_R\big]^\top \cdot\frac{1}{|V_r^t|}\sum_{v'\in V_r^t} \big[\tanh\big(\mathbf{W}_R^{t,l}\cdot\mathbf{h}^{t,l}_{v',r}+\mathbf{b}\big)\big],
\end{equation}
where $V_r^t$ denotes the set of nodes connected by relation $r$ at timestamp $t$, $\mathbf{b}\in \mathbb{R}^{d}$ is the bias vector, $\mathbf{W}_{R}^{t,l}\in \mathbb{R}^{d\times d}$ and $\mathbf{c}_{R}^{t,l}\in \mathbb{R}^{d}$ are trainable transformation matrix and attention vector, respectively. The normalized importance of $r$ with regard to $v$ is computed as:
\begin{equation}
	\beta_{v,r}^{t,l} = \frac{\exp\Big(e^{t,l}_{r}\Big)}{\sum_{r' \in R(v)} \exp\Big(e^{t,l}_{r'}\Big)}.
\end{equation}
With the importance of different relations, we generate the spatial embedding for $v$ via linear combination:
\begin{equation}
	\mathbf{h}^{t,l}_{v,R} = \sum_{r\in R(v)}\big[\beta_{v,r}^{t,l}\big] \cdot \big[\mathbf{h}^{t,l}_{v,r}\big].
\end{equation}
An intuitive explanation is shown in Figure \ref{fig:sys} (d). We could also extend it to multi-head mechanism.

\subsection{Across-time Aggregation}

Intra-relation and inter-relation aggregation modules that aggregate information from the target node's spatial neighbors are performed purely on each graph slice. Across-time aggregation aims to capture the interactions among the target node's temporal neighbors. We define the temporal neighbors as the same nodes in different graph slices (including itself). This module takes in the spatial embeddings of the target node's temporal neighbors and outputs a spatial-temporal embedding for this target node. This process is formalized as follow: 
\begin{equation}
	\mathbf{h}^{t,l}_{v,ST}=AGG_{across}\Big(\big\{\mathbf{h}^{t',l}_{v,R}|1\leq t' \leq T \big\}; \boldsymbol{\Theta}_{across}\Big),
\end{equation}
where $\mathbf{h}^{t',l}_{v,R}$ is the spatial embedding of node $v$'s temporal neighbor at timestamp $t'$ in layer $l$, $\mathbf{h}^{t,l}_{v,ST}$ is node $v$'s spatial-temporal embedding at timestamp $t$ in layer $l$, $\boldsymbol{\Theta}_{across}$ is the trainable parameters and is non-shareable for node type and aggregation layer.

As the transformer \cite{vaswani2017attention} has shown great performance in natural language processing domain, we explore its attention mechanism to model our across-time aggregation process. Before calculating attentions for different timestamps, we define a time encoding function $PE(\cdot)$ for $\mathbf{h}^{t,l}_{v,R}$ to incorporate time-related factors:
\begin{equation}
	PE\Big(\mathbf{h}^{t,l}_{v,R}\Big) = \|_{i=1}^{d}\Big(\mathbf{h}^{t,l}_{v,R}(i)+p(t, i)\Big),
\end{equation}
where $i$ is the index of each element and $p(\cdot)$ is a frequency encoding function that characterizes a time-dependent sinusoid, where $p(t,i) = \sin(t/10000^{2i/d})$ if $i$ is even, or $\cos(t/10000^{2i/d})$ if odd. By feeding the embeddings at different timestamps into this function, they become discriminative with regard to time. We then transform the target node's spatial embedding into Query vector, its temporal neighbor's spatial embedding into Key vector, and calculate their dot product as the attention coefficient to measure the importance of this temporal neighbor. Accordingly, we have: 
\begin{equation}
	\begin{aligned}
		\mathbf{q}^{t,l}_{v}  & = \mathbf{W}^{l}_{\phi(v),q} \cdot PE\Big(\mathbf{h}^{t,l}_{v,R}\Big),\\
		\mathbf{k}^{t',l}_{v} & = \mathbf{W}^{l}_{\phi(v),k} \cdot PE\Big(\mathbf{h}^{t',l}_{v,R}\Big),
	\end{aligned}
\end{equation}
where $\mathbf{W}^{l}_{\phi(v),q}, \mathbf{W}^{l}_{\phi(v),k} \in \mathbb{R}^{d\times d}$ denote the trainable transformation matrices for query, and key, respectively. The normalized attention value is then calculated by: 
\begin{equation}
	\gamma^{t',l}_v = \frac{\exp\Big(\big[\mathbf{q}^{t,l}_{v}\big]^\top \cdot \big[\mathbf{k}^{t',l}_{v}\big]\Big)}{\sum_{t''=1}^T \exp\Big(\big[\mathbf{q}^{t,l}_{v}\big]^\top \cdot \big[\mathbf{k}^{t'',l}_{v}\big]\Big)}.
\end{equation}
Finally, node $v$'s spatial-temporal embedding is computed via a linear combination of its temporal neighbors' transformed embeddings and the calculated attention values, formulated as: 
\begin{equation}
	\begin{aligned}
		\mathbf{v}^{t',l}_{v} & = \mathbf{W}^{l}_{\phi(v),v} \cdot PE\Big(\mathbf{h}^{t',l}_{v,R}\Big),\\
		\mathbf{h}^{t,l}_{v,ST} &= \sigma\Big(\sum_{t'=1}^T \big[\gamma^{t',l}_v\big] \cdot \big[\mathbf{v}^{t',l}_{v}\big]\Big).
	\end{aligned}
\end{equation}
A graphical explanation of across-time aggregation is shown in Figure \ref{fig:sys} (e). Similarly, multi-head attention mechanism could also be applied in this module.

Using the hierarchical aggregation mechanism above, we can obtain the spatial-temporal embedding for each node. Despite this, the feature vector of node itself also plays an essential role in learning its representation. Instead of directly summing them, we design a gate mechanism to control how much information the node and its neighbors should contribute. Given the node feature vector at timestamp $t$ in layer $l-1$ and its spatial-temporal embedding at the same timestamp in layer $l$, their combination is formulated as:
\begin{equation}
	\mathbf{h}^{t,l}_{v} = \delta_{\phi(v)}\cdot \big[\mathbf{h}^{t,l}_{v,ST}\big]+(1-\delta_{\phi(v)}) \cdot \big[\mathbf{W}_{\phi(v)} \cdot \mathbf{h}^{t,l-1}_{v}\big],
\end{equation}
where $\delta_{\phi(v)} \in \mathbb{R}^{1}$ and $\mathbf{W}_{\phi(v)} \in \mathbb{R}^{d\times d}$ are the trainable weight and transformation matrix, respectively.


\begin{table*}[!h]
	\centering
	\caption{Statistics of datasets.}
	\vspace{0.1cm}
	\label{tab:data}
	\scalebox{0.78}{
	\begin{tabular}{|c|c|c|c|c|c|c|}
		\hline
		Dataset &
		Graph &
		Time Span &
		Node &
		Relation &
		Feature &
		Data Split \\ \hline
		OGBN-MAG &
		\begin{tabular}[c]{@{}c@{}}\# Graph: 10\\ Granularity: year\end{tabular} &
		\begin{tabular}[c]{@{}c@{}} 2010-2019\end{tabular} &
		\begin{tabular}[c]{@{}c@{}}\# Author: 17,764\\ \# Paper: 282,039\\ \# Field: 34,601\\ \# Institution: 2,276\end{tabular} &
		\begin{tabular}[c]{@{}c@{}}\# Author-Paper: 2,061,677\\ \# Paper-Paper:2,377,564\\ \# Paper-Field: 289,376\\ \# Author-Institution: 40,307\end{tabular} &
		\begin{tabular}[c]{@{}c@{}}Paper: 128\\ Author: 128\\ Institution: 128\\ Field: 128\end{tabular} &
		\begin{tabular}[c]{@{}c@{}}Training: 8\\ Validation: 1\\ Testing: 1\end{tabular} \\ \hline
		COVID-19 &
		\begin{tabular}[c]{@{}c@{}}\# Graph: 304\\ Granularity: day\end{tabular} &
		\begin{tabular}[c]{@{}c@{}} 05/01/2020-\\  02/28/2021\end{tabular} &
		\begin{tabular}[c]{@{}c@{}}\# State: 54\\ \# County: 3223\end{tabular} &
		\begin{tabular}[c]{@{}c@{}}\# State-State: 269 \\ \# State-County: 3,141\\ \# County-County: 22,176 \end{tabular} &
		\begin{tabular}[c]{@{}c@{}}State: 1\\ County: 1\end{tabular} &
		\begin{tabular}[c]{@{}c@{}}Training: 244\\ Validation: 30\\ Testing: 30\end{tabular} \\ \hline
	\end{tabular}
	}\vspace{-0.3cm}
\end{table*}

\subsection{Learning Algorithm}

By stacking $L$ heterogeneous temporal aggregation layers, we could derive the embedding for each node at each timestamp, denoted as $\mathbf{h}_v^{t,L}$. We then simply sum the node embedding of all timestamps as its final embedding: $\mathbf{h}_v = \sum_{t=1}^{T}\mathbf{h}_v^{t,L}$. HTGNN could be trained in an end-to-end manner with the labeled data at timestamp $T+1$, as following:
\begin{equation}\label{eq:loss}
	\mathcal{L} = \sum_{v\in \mathcal{V}_L} J(y_v, \hat{y}_v) +\lambda\|\boldsymbol{\Theta}\|_2^2,\hat{y}_v   = \sigma(MLP(\mathbf{h}_v)).
\end{equation}
where $J(\cdot)$ measures the loss between ground $y_v$ and the predicted score $\hat{y}_v$, $\|\boldsymbol{\Theta}\|_2^2$ is the L2-regularizer to prevent over-fitting. Depending on the goals of different tasks, $J(\cdot)$ could be set as cross-entropy loss for node classification and link prediction problems, or mean absolute error for regression problem.

\section{Experiments}\label{sec:exp}

In this section, we conduct four sets of experiments to evaluate the performance of the proposed HTGNN.

\subsection{Datasets}

We construct two HTGs from two different domains with distinct characteristics. OGBN-MAG dataset, an academic network with dynamically evolving heterogeneous structures, is used for link prediction task. COVID-19 dataset, an epidemiological network with constantly changing node features, is used for node regression task. 

\noindent \textbf{OGBN-MAG:} The original OGBN-MAG dataset \cite{hu2020open} is a static heterogeneous network composed of a subset of the Microsoft Academic Graph (MAG). We extract a HTG from OGBN-MAG consisting of 10 graph slices spanning from 2010 to 2019. We first select authors that consecutively publish at least one paper every year. We further collect these authors' affiliated institutions, published papers, and the papers' field of studies in each year to construct this HTG. Each graph slice is a heterogeneous graph that contains four types of nodes (paper, author, institutions, and fields of study), and four types of relations among them (author-\textit{affiliated with}-institution, author-\textit{writes}-paper, paper-\textit{cites}-paper, and paper-\textit{has a topic of}-field of study).

\noindent \textbf{COVID-19:} The COVID-19 data is obtained from 1point3acres\footnote{https://coronavirus.1point3acres.com/en}, which contains both state and county level daily case reports (e.g., confirmed cases, new cases, deaths, and recovered cases). We use the daily new COVID-19 cases as the time-series data for each state and county. We then build a HTG including 304 graph slices spanning from 05/01/2020 to 02/28/2021. Each graph slice is also a heterogeneous graph consisting of two types of nodes (state and county) and three types of relations between them, i.e., one administrative affiliation relation (state-\textit{includes}-county) and two geospatial relations (state-\textit{near}-state, county-\textit{near}-county).

In OGBN-MAG, each paper comes with a 128-dimensional feature vector obtained by averaging the embeddings of words in its title and abstract. For other nodes, we run the metapath2vec \cite{dong2017metapath2vec} on each graph slice to generate the 128-dimensional node embeddings as their input features. For COVID-19, we attach each node in each graph slice with its daily new cases as the node feature. We split each dataset into training, validation, and testing sets with a ratio of 8:1:1. Statistics of these datasets are summarized in Table \ref{tab:data}.

\begin{table*}[!h]
	\centering
	\caption{Experimental results of different methods on OGBN-MAG (\%) and COVID-19 datasets.}
	\label{tab:baseline}
    \scalebox{0.78}{
	\begin{tabular}{|c|c|c|c|c|c|c|c|c|c|c|c|c|c|}
		\hline
		\multirow{2}{*}{Dataset} &
		\multicolumn{2}{c|}{\multirow{2}{*}{Metric}} &
		\multicolumn{2}{c|}{Sequence Model} &
		\multicolumn{4}{c|}{Static Graph Model} &
		\multicolumn{4}{c|}{Dynamic Graph Model} &
		Ours \\ \cline{4-14} 
		& \multicolumn{2}{c|}{}                                 & LSTM  & TRFM  & GCN   & GAT   & RGCN  & HGT   & CoGNN & DySAT & HDGAN & DyHATR    & HTGNN  \\ \hline
		\multirow{4}{*}{OGBN-MAG} &\multirow{2}{*}{AUC} & Mean  & 81.37 & 82.89 & 78.81 & 80.23 & 80.34 & 85.30 & 85.87 & 86.36 & 88.88 & 89.49     & \textbf{91.01} \\ 
		&                                               & SD    & 0.42  & 0.36  & 1.53  & 2.07  & 2.21  & 1.20  & 1.52  & 0.24  & 0.73  & 0.65      & 0.77 \\ \cline{2-14} 
		& \multirow{2}{*}{AP}                           & Mean  & 78.56 & 79.81 & 76.46 & 77.58 & 78.11 & 82.68 & 83.11 & 83.83 & 86.18 & 86.24     & \textbf{89.18} \\ 
		&                                               & SD    & 0.29  & 0.25  & 1.95  & 1.74  & 1.65  & 1.20  & 0.92  & 0.29  & 0.82  & 0.91      & 1.24 \\ \hline
		\multirow{4}{*}{COVID-19} & \multirow{2}{*}{MAE}& Mean  & 720   & 691   & 846   & 821   & 833   & 805   & 653   & 672   & 656   & 643       & \textbf{555}    \\  
		&                                               & SD    & 115   & 54    & 101   & 91    & 95    & 88    & 45    & 74    & 66    & 36        & 34  \\ \cline{2-14} 
		& \multirow{2}{*}{RMSE}                         & Mean  & 1504  & 1458  & 1674  & 1612  & 1640  & 1598  & 1298  & 1373  & 1303  & 1282      & \textbf{1136}   \\ 
		&                                               & SD    & 258   & 182   & 204   & 211   & 198   & 200   & 60   & 96    & 81    & 59        & 65     \\ \hline
	\end{tabular}
	}\vspace{-0.4cm}
\end{table*}

\vspace{-0.3cm}
\subsection{Baselines}

We compare HTGNN with three classes of state-of-the-art baselines.

\noindent \textbf{Neural Sequence Models}: This class of baselines is capable of capturing temporal dependencies. LSTM \cite{hochreiter1997long} is a type of recurrent neural network that learns order dependence of sequences. Transformer \cite{vaswani2017attention} handles sequences with global message routing, weighing the influence of different parts of the input.

\noindent \textbf{Static Graph Models}: We consider several static homogeneous/heterogeneous GNNs that depict spatial dependencies. GCN \cite{kipf2016semi} and GAT \cite{velivckovic2017graph} work on homogeneous graphs, where the neighbor information is aggregated through a mean function and a self-attention mechanism, respectively. For the heterogeneous GNNs, we choose RGCN \cite{schlichtkrull2018modeling} and HGT \cite{hu2020heterogeneous} that do not rely on metapaths. RGCN considers specialized transformation matrices for different types of relations. HGT applies the Transformer architecture to learn the mutual attention for each meta relation. In the experiment, we treat HGT as a static heterogeneous GNN as the relative temporal encoding is not applicable for HTGs.

\noindent \textbf{Dynamic Graph Models}: We select one spatial-temporal GNN, one dynamic homogeneous GNN, and two dynamic heterogeneous GNNs as baselines. CoGNN \cite{kapoor2020examining} applies a multilayer perceptron to process time-series node features and uses GCN \cite{kipf2016semi} with skip connections for spatial information aggregation. DySAT \cite{sankar2020dysat} employs self-attention to aggregate structural neighborhood and temporal dynamics for node representation learning. HDGAN \cite{ji2021dynamic} combines heterogeneous attention and Hawkes process to model graph heterogeneity and dynamics. We replace the heterogeneous attention module with HGT \cite{hu2020heterogeneous} to avoid incorporating metapaths. DyHATR \cite{xue2020modeling} uses hierarchical attention to learn heterogeneous information and incorporates RNNs with temporal attention to capture temporal dependencies.


\subsection{Implementation Details}
For those models designed for homogeneous graphs, we ignore the graph heterogeneity and directly feed the whole graph into the learning algorithms. We employ Adam optimizer with learning rate set to 5e-3, and weight decay set to 5e-4. For other parameters, we set dropout rate to 0.2, GNN layer to 2, hidden embedding dimension to 32 for OGBN-MAG and 8 for COVID-19, respectively; and we also use ReLU as the activation function. We train all the models with a fixed 500 epochs and use an early stopping strategy with a patience of 50. That is to say, the best models are selected when the validation loss does not decrease for 50 consecutive epochs. All models are trained for five times, and the mean and standard deviation of test performance are reported. All baselines and the proposed HTGNN are implemented with Python 3.7.10, PyTorch 1.8.1 and Deep Graph Library (DGL) 0.6.0. Experiments are conducted on a machine equipped with i9-9900K processor, two RTX 2080Ti graphic cards, and 64 GB of RAM.

\subsection{Link Prediction}

We conduct the link prediction experiment on the OGBN-MAG dataset to evaluate the performances of different methods. We split the dataset into three sets with a ratio of 8:1:1. Specifically, the graphs of 2010-2017 are used for training, the graph of 2018 for validation, and the graph of 2019 for testing. We frame our task to new co-author link prediction. The new co-author relation is defined as the co-author link that exists in year $T+1$ but not in year $T$. We randomly select 10\% new co-author links as positive samples. Following the standard manner of learning-based link prediction, we randomly sample the same number of nonexistent co-author links as negative samples. We set the time window size to 3, which means, to predict the co-author relation in next year, we consider the HTG of the past three years. Note that, for static graph models without considering temporal dependency, we simply set the time window to 1. For a pair of authors, after obtaining the embeddings via HTGNN, we feed their concatenation into Eq. (\ref{eq:loss}) for training with the cross-entropy loss. Similar to \cite{zhang2018link, fan2021heterogeneous}, we adopt the widely used AUC score (the Area Under a receiver operating characteristic Curve) and AP score (Average Precision) to measure the co-author link prediction performance.

The experimental results with mean performance and their standard deviations reported are shown in Table \ref{tab:baseline}. We have the following conclusions by analyzing the results: (1) Both sequence and static graph models could achieve satisfactory results, which indicates that the temporal and spatial dependencies depicted by these two types of methods contribute to the co-author link prediction problem. (2) Dynamic graph models improve the performance by taking the information in both spatial and temporal domains into consideration. (3) GNNs designed for heterogeneous graphs (i.e., RGCN, HGT, HDGAN, DyHATR) perform better than homogeneous GNNs (i.e., GCN, GAT, CoGNN, DySAT), which demonstrates the advantage of incorporating graph heterogeneity. (4) Our proposed HTGNN that could jointly model the heterogeneous spatial dependencies and temporal dimensions consistently outperforms all baselines.

\subsection{Node Regression}

The node regression task is conducted on the COVID-19 dataset. We aim to perform state-level daily new case forecasting. We also split the dataset into training, validation, and testing sets with a ratio of 8:1:1. Specifically, 05/01/2021-12/30/2020 are used for training, 12/31/2020-01/29/2021 for validation, and 01/30/2021-02/28/2021 for testing. In this task, we set the time window to 7 (using the past one-week historical data for forecasting). Similarly, we set it to 1 for static graph models. As suggested in \cite{yu2017spatio,deng2020cola}, MAE (Mean Absolute Errors) and RMSE (Root Mean Squared Errors) are adopted to measure the performance. We report the average MAE and RMSE of 54 states in the US. The experimental results with mean and standard deviation reported are demonstrated in Table \ref{tab:baseline}. Besides some similar conclusions drawn from the link prediction task, we notice that sequence models yield better performances than static graph models. This phenomenon indicates that the temporal domain contributes more to the COVID-19 forecasting task compared to the spatial domain. This is because the graph structures remain unchanged in this case; however, the node features (i.e., daily new cases) have a remarkable change over time.




\subsection{Ablation Study}

In HTGNN, we propose to model heterogeneous spatial dependencies and temporal dimensions jointly. To evaluate this design, we establish two variants of HTGNN for comparison that handle these two types of dependencies serially. HTGNN$_{ST}$ processes spatial dependencies first and temporal dependencies later by first performing multi-layer intra- and inter-relation aggregations in each graph slice, then applying across-time aggregation across all graph slices. HTGNN$_{TS}$ analyzes these two domains in reverse order by first employing across-time aggregation on the temporal domain and then conducting multi-layer spatial aggregation on the last graph slice. Experimental results shown in Table \ref{tab:sp} demonstrate that: (1) HTGNN$_{ST}$ outperforms HTGNN$_{TS}$ on the OGBN-MAG dataset but is less effective on the Covid-19 dataset. This is because HTGNN$_{ST}$ emphasizing more on spatial domain fits better with OGBN-MAG with dynamically evolving graph structures. On the contrary, HTGNN$_{TS}$ paying more attention to the temporal domain is more suitable for the COVID-19 with changing node features. (2) HTGNN achieves better performance than these two variants in both datasets, proving that our proposed holistic model is agnostic to graph characteristics and delivers superior performance. 

\begin{figure}[!h]
	\centering
	\includegraphics[width=1\linewidth]{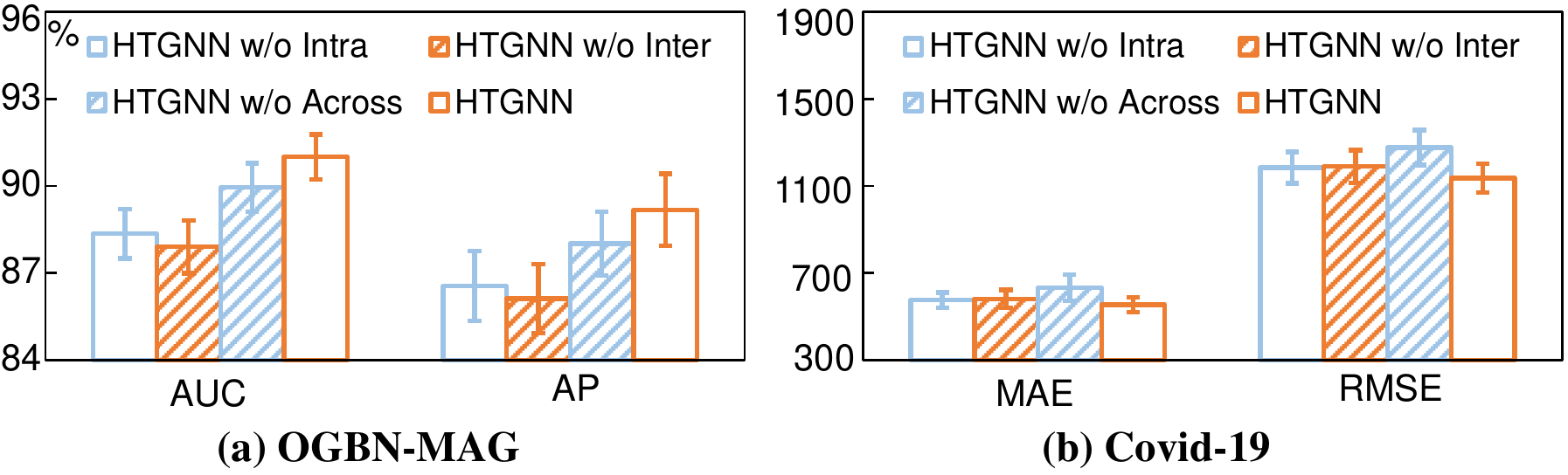}\vspace{-0.2cm}
	\caption{Evaluation of each component in  HTGNN.}\vspace{-0.5cm}
	\label{fig:ablation}
\end{figure}

We then perform additional ablation studies to evaluate the three major components in HTGNN: intra-relation, inter-relation, and across-time aggregation modules. Accordingly, we prepare three variants to examine the effect of each component. HTGNN w/o Intra replaces the intra-relation aggregation module in each layer with a mean pooling mechanism. HTGNN w/o Inter replaces the inter-relation aggregation module in each layer with a mean pooling mechanism. HTGNN w/o Across replaces the across-time aggregation module in each layer with a mean pooling mechanism. Experimental results are shown in Figure \ref{fig:ablation}. From Figure \ref{fig:ablation}, we observe that HTGNN equipped with three components achieves the best performance, which proves that each component makes its contribution to the final performance. It is also worth noting that HTGNN w/o Intra and HTGNN w/o Inter work better than HTGNN w/o across on the COVID-19 dataset but yield worse results on the OGBN-MAG dataset. We owe this to the distinct characteristics of different datasets. In particular, for OGBN-MAG, the heterogeneous graph structures evolve dynamically, increasing the difficulty in capturing spatial dependencies. In contrast, for COVID-19, the temporal dependencies are relatively harder to capture as the node features change constantly, while the graph structures remain unchanged.

\begin{table}[!h]
\vspace{-0.3cm}
\centering
\caption{Evaluation of joint modeling on HTGs.}
\label{tab:sp}
\scalebox{0.75}{
\begin{tabular}{|c|c|c|c|c|}
\hline
Dataset & \multicolumn{2}{c|}{OGBN-MAG} & \multicolumn{2}{c|}{COVID-19} \\
\hline
Metric  & AUC           & AP           & MAE          & RMSE          \\
\hline
HTGNN$_{ST}$      & $89.35\pm0.81$  & $86.62\pm0.78$ & $640\pm50$   & $1270\pm75$          \\
HTGNN$_{TS}$      & $88.01\pm0.71$  & $85.11\pm1.04$ & $619\pm38$   & $1232\pm81$          \\
\hline
HTGNN   & $\textbf{91.01}\pm0.77$  & $\textbf{89.18}\pm1.24$ & $\textbf{555}\pm34$   & $\textbf{1136}\pm65$         \\
\hline
\end{tabular}
}\vspace{-0.3cm}
\end{table}

\subsection{Parameter Sensitivity}

In this section, we investigate HTGNN's sensitivity to key hyper-parameters.

\begin{figure}[h]
	\centering
	\includegraphics[width=1\linewidth]{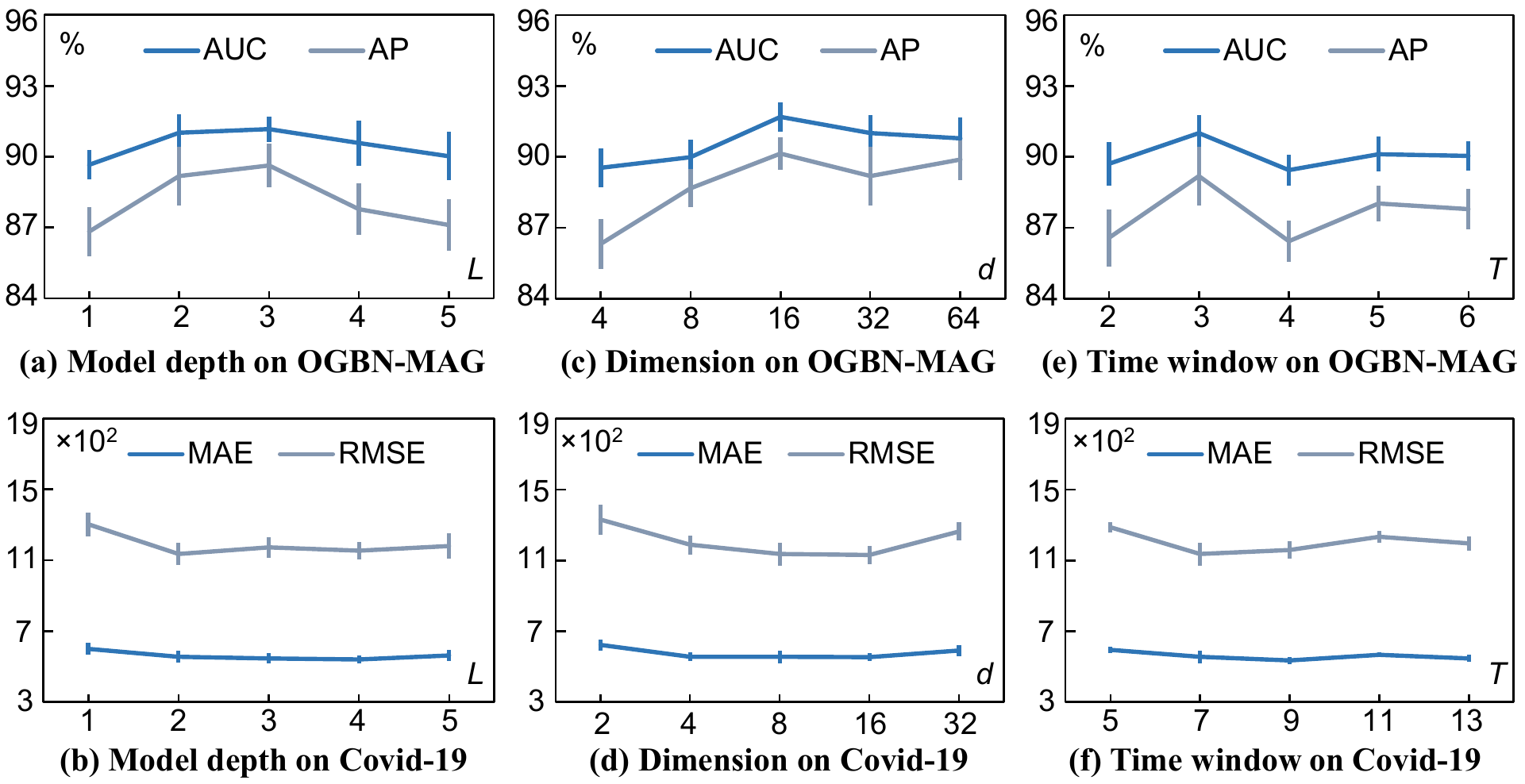}
	\caption{Parameter sensitivity analysis.}
	\label{fig:para}
\end{figure}

\noindent\textbf{Model depth.} We vary the model depth (i.e., the number of heterogeneous temporal aggregation layers) from 1 to 5 to examine the model's performance on two datasets. The experimental results with mean and standard deviation reported are shown in Figure \ref{fig:para} (a)-(b). We can see that with the increase of model depth, the performance of HTGNN first improves then starts to decrease gradually. This phenomenon is attributed to the oversmoothing problem.  

\noindent\textbf{Embedding dimension.}  We vary the embedding dimension from 4 to 64 for OGBN-MAG and 2 to 32 for COVID-19 to investigate its influence. Comparison results are illustrated in Figure \ref{fig:para} (c)-(d). We observe that increasing the embedding dimension initially improves the performance since a larger dimension can preserve more information. However, when using a too large dimension, the model would suffer the overfitting problem, which results in reduced performance. 

\noindent\textbf{Time window size.} We validate the effect of time window size by ranging it from 2 to 6 for OGBN-MAG and 5 to 13 for COVID-19, respectively. The results are shown in Figure \ref{fig:para} (e)-(f). We can see that a large time window size boosts the performance as more historical information is included. However, further enlarging the window size yields a fluctuating performance.


\section{Conclusion}\label{sec:conclusion}

In this paper, we study the representation learning problem on heterogeneous temporal graphs (HTGs), a general concept for modeling heterogeneous and constantly evolving graph data. We further propose heterogeneous temporal graph neural network (HTGNN), a holistic framework tailored heterogeneity with evolution in time and space for HTG representation learning. In particular, HTGNN consists of several heterogeneous temporal aggregation layers, each of which employs a hierarchical aggregation mechanism, including intra-relation, inter-relation and across-time aggregation modules, to jointly model heterogeneous spatial dependencies and temporal dimensions. Extensive experiments are conducted on two built HTGs: OGBN-MAG with dynamically evolving heterogeneous structures, and COVID-19 with constantly changing node features. Promising results demonstrate the great performance of HTGNN in comparison with state-of-the-art baselines.

\balance
\bibliographystyle{IEEEtran}
\bibliography{ref}

\end{document}